\documentclass[runningheads]{llncs}
\usepackage[T1]{fontenc}
\usepackage{amsmath}
\usepackage[ruled,vlined]{algorithm2e}
\SetAlgoCaptionSeparator{} % Removes the colon after "Algorithm X"
\SetEndCharOfAlgoLine{}
\SetKwInput{KwInput}{Input}
\SetKwInput{KwOutput}{Output}

\newcommand{\cossim}{\operatorname{cos\,sim}}
\newcommand{\hhfill}{\hfill\hfill\hfill\hfill\hfill\hfill\hfill\hfill\hfill}

\usepackage{hyperref}
\usepackage{url}
\usepackage{graphicx}
\usepackage{subcaption}
\usepackage{booktabs}

\usepackage{amsmath}
\usepackage{amssymb}
\usepackage{amsfonts}

\begin{document}
\title{Normalized Matching Transformer}
\author{Abtin Pourhadi \and
Paul Swoboda}
\authorrunning{A. Pourhadi and P. Swoboda}
\institute{Heinrich Heine University D\"usseldorf, Germany
\email{\{abtin.pourhadi,paul.swoboda\}@hhu.de}}
\maketitle              % typeset the header of the contribution
\begin{abstract}
We introduce the \textbf{Normalized Matching Transformer (NMT)}, a deep learning approach for efficient and accurate sparse semantic keypoint matching between image pairs.
NMT consists of a strong visual backbone, geometric feature refinement via SplineCNN, followed by a normalized Transformer for computing matching features.
Central to NMT is our \emph{hyperspherical normalization strategy}: we enforce unit‐norm embeddings at every Transformer layer and train with a combined contrastive InfoNCE and hyperspherical uniformity loss to yield more discriminative keypoint representations.
This novel architecture/loss combination encourages close alignment of matching image features and large distances between non-matching ones not only at the output level, but for each layer. Despite its architectural simplicity, NMT sets a new state-of-the-art performance on PascalVOC and SPair-71k, outperforming BBGM~\cite{rolinek2020deep}, ASAR~\cite{ren2022appearance}, COMMON~\cite{lin2023graph} and GMTR~\cite{guo2024gmtr} by 5.1\% and 2.2\%, respectively, while converging in at least $1.7\times$ fewer epochs compared to other state-of-the-art baselines.
These results underscore the power of combining pervasive normalization with hyperspherical learning for matching tasks.
\end{abstract}
\section{Introduction}
\label{sec:intro}
Traditional graph-matching pipelines~\cite{rolinek2020deep,torresani2012dual} rely on neural network backbones for computing discriminative features combined with combinatorial solvers to establish keypoint correspondences to address the feature matching problem.
While effective, these approaches are often complex, requiring the combination of a neural network based feature computation stage with an intricate combinatorial stage for computing keypoint correspondences.
Integrating the combinatorial stage into a neural network pipeline brings its own challenges, including non-differentiability and most often combinatorial solvers running on CPUs.
Recent methods like GMTR~\cite{guo2024gmtr}, ASAR~\cite{ren2022appearance} and COMMON~\cite{lin2023graph} proposed pure deep-learning approaches with a simpler Sinkhorn-based decoding.
They have sought to enhance performance and robustness through Transformer-based architectures, better losses and/or specialized regularization strategies.
These newer approaches, while foregoing a combinatorial stage, still outperform hybrid approaches, attesting to the strength of deep learning even in the setting of keypoint matching that has a strong combinatorial aspect to it.

While pure deep learning methods have already reached very high results on semantic keypoint matching datasets, we show that there is still room for improvement.
First, better backbones boost performance. We replace the commonly used VGG~\cite{simonyan2014very} backbone with a Swin Transformer~\cite{liu2021swin}.
Second, we process features at keypoints with a SplineCNN~\cite{fey2018splinecnn}, which adds helpful inductive biases incorporating the geometry of the keypoints to match.
Third, we use Transformers to exchange information within and across images. We show that vanilla Transformers can be outperformed by additional normalization techniques used in normalized Transformers~\cite{loshchilov2024ngpt}.
We argue that the employed normalization techniques are well-suited for our normalized feature representation: Instead of only normalizing at the end before computing cosine similarity and the losses, we normalize throughout the normalized Transformer, which helps in faster training and better overall performance.
Our pipeline is trained using a contrastive~\cite{oord2018representation} and hyperspherical~\cite{mettes2019hyperspherical} loss together with data augmentation.

Our work shows that the performance of semantic keypoint matching, one of the classical and well-studied problems in computer vision, has not yet saturated and can be enhanced by leveraging current deep learning methods.
In particular, we argue that our contribution of hyperspherical architecture and losses enhances feature quality and improves training speed.
Also, no combinatorial subroutines are necessary given the capabilities of our neural network pipeline, simplifying our overall approach.

\begin{figure*}[!t]
    \centering
    \includegraphics[width=0.95\textwidth]{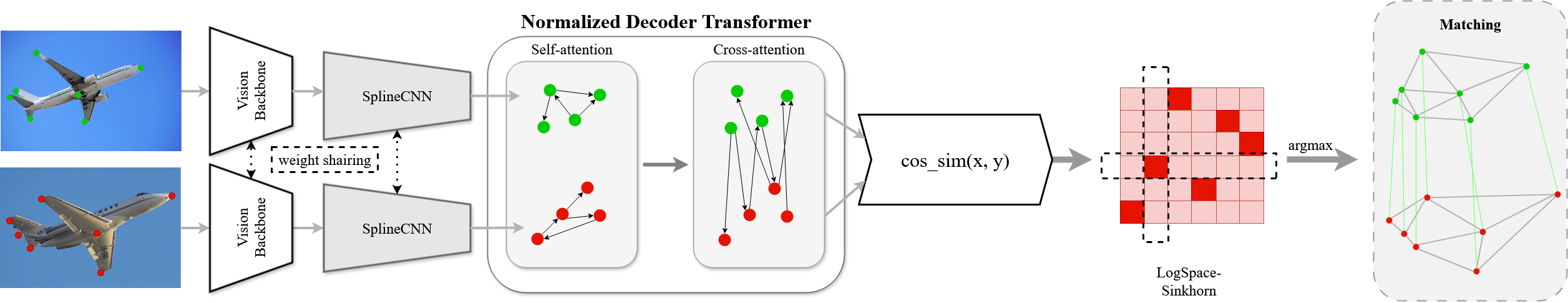}
    \caption{
    Normalized matching Transformer inference.
    Each image in a pair is passed through a Swin Transformer visual backbone.
    Features at keypoints are extracted and given through a SplineCNN for further feature refinement.
    Two normalized Transformer decoders interleave self-attention between keypoint features from the same image with cross-attention that mixes information across images.
    Finally, cosine similarities are computed and given as affinities to a log-space Sinkhorn routine from which a matching is decoded.
    }
    \label{fig:matching-pipeline}
\end{figure*}
\paragraph{Contribution.}
In detail, our contributions are as follows:
\begin{description}
    \item[\textbf{Architecture}:] We propose a simple and efficient pure ML-based architecture combining
    an image processing backbone using a Swin Transformer~\cite{liu2021swin}, followed by a keypoint feature processing stage consisting of SplineCNN~\cite{fey2018splinecnn}, a graph neural network that exploits the geometrical structure of keypoints.
    \item[\textbf{Decoding}:]
    The feature computation stage is followed by a two‐stream Transformer decoder, each stream processing keypoints from one image. Each decoder employs normalized Transformer layers~\cite{loshchilov2024ngpt} with unit‐norm parameter normalization, decoupled attention and MLP residual pathways to stabilize gradient flow throughout the network. Raw cosine‐similarity affinities are computed via the normalized decoder outputs. Only during inference, a differentiable Sinkhorn algorithm~\cite{cuturi2013sinkhorn} converts these affinities into a soft, doubly stochastic matching matrix. This design removes the need for combinatorial solvers and yields a streamlined, end‐to‐end matching pipeline.
    \item[\textbf{Loss}:] Our method incorporates improved loss formulations, including InfoNCE~\cite{oord2018representation} and hyperspherical loss~\cite{mettes2019hyperspherical}.
    They improve feature embedding quality and ensure robust training, enabling the model to learn more discriminative representations.
    \item[\textbf{Experimental}:] Extensive experiments demonstrate state-of-the-art performance on PascalVOC and SPair-71k datasets, with significant improvements over the state-of-the-art methods BBGM~\cite{rolinek2020deep}, ASAR~\cite{ren2022appearance}, COMMON~\cite{lin2023graph} and GMTR~\cite{guo2024gmtr} exceeding their performance by 5.1\% on PascalVOC and 2.2\% on SPair-71k.
    We also need at least $1.7\times$ fewer epochs until training convergence than the baselines.
\end{description}

\section{Related work}

\begin{figure*}[t]
    \begin{center}
    \includegraphics[width=0.95\linewidth]{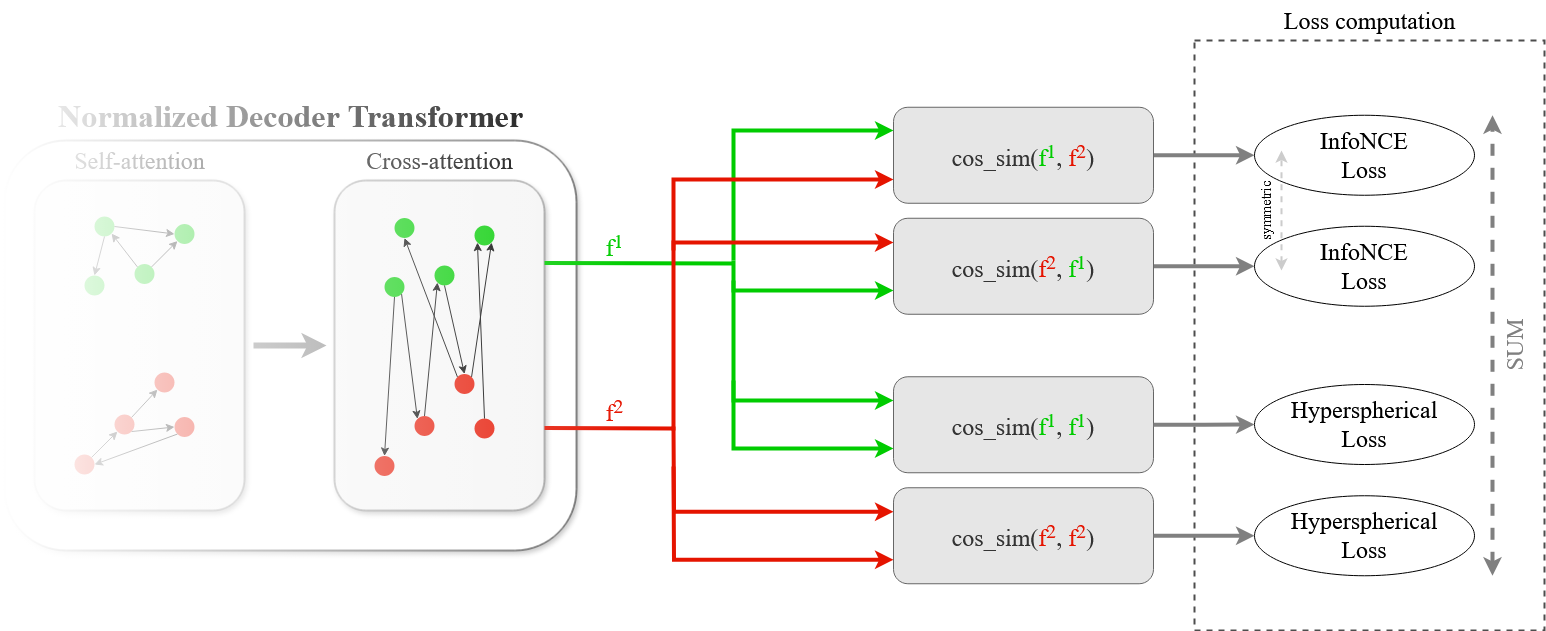}
    \caption{
    Normalized matching Transformer losses.
    Losses are applied on the features that are computed by the normalized Transformer decoder.
    InfoNCE losses are computed on cosine similarities of features coming from a single keypoint in one image and all keypoint features from the other one and align matching correspondences.
    For symmetry we apply the InfoNCE loss in both directions.
    For distributing features of different keypoints in the same image we use a hyperspherical loss on the keypoint features coming from each image separately.
    }
    \label{fig:losses-pipeline}
    \end{center}
\end{figure*}

\label{sec:relatedWork}
Related work on keypoint matching involves
(i)~\emph{combinatorial} aspects for establishing a one-to-one correspondence between sets of keypoints,
(ii)~\emph{hybrid} approaches that combine mainly neural networks for computing keypoint features with combinatorial routines to get correspondences and
(iii)~\emph{pure deep learning} based methods that forego any combinatorial subroutines.

On the application side we distinguish between (i)~\emph{sparse semantic} keypoint matching, which we study here, for computing correspondences between few select keypoints of distinct objects of the same class in different environments and (ii)~\emph{dense geometric} keypoint matching for estimating homographies between many keypoints belonging to the same object in the same scene but e.g.\ viewed from different viewpoints.

\paragraph{Combinatorial Aspects \& Assignment Problems.}
In the combinatorial literature, establishing one-to-one correspondences is framed as an assignment problem. Linear assignment, relying only on unary costs is polynomially solvable and often approximated in deep learning via the Sinkhorn algorithm~\cite{cuturi2013sinkhorn}. Conversely, the quadratic assignment problem (graph matching) incorporates pairwise geometric costs but is notoriously NP-hard~\cite{lawler1963quadratic,burkard1997qaplib}. While advanced combinatorial solvers exist~\cite{torresani2012dual,kahl2024unlocking}, they remain computationally expensive and difficult to integrate seamlessly into end-to-end differentiable pipelines.

\paragraph{Hybrid Approaches.}
For the keypoint matching problem the traditional approach is to first extract discriminative features for each keypoint (resp.\ for pairs of keypoints), use those to compute costs for matching keypoints and finally to compute correspondences using the linear or quadratic assignment problem.
Some approaches use ad-hoc heuristics for decoding correspondences, e.g. via reformulation to constrained vertex classification~\cite{wang2021neural} or the quadratic assignment problem~\cite{torresani2012dual,rolinek2020deep,wang2021neural}.

Pre-neural network approaches with hand-crafted feature descriptors were for quite some time still state of the art~\cite{torresani2012dual}.
However, neural network features eventually overtook~\cite{zanfir2018deep}.
Follow-up work NGM~\cite{wang2021neural} differentiates through the construction of a quadratic assignment problem and decodes the matching by converting to a constrained vertex classification problem.
The hybrid approach~\cite{rolinek2020deep} combined a state-of-the-art neural network pipeline with a quadratic assignment solver and used a special backpropagation technique~\cite{vlastelica2019differentiation} to learn in tandem with the non-differentiable combinatorial solver.

\paragraph{Pure Deep-Learning.}
When not using combinatorial routines it is even more important to obtain discriminative features.
One of the first pure neural network methods~\cite{zanfir2018deep} relaxed a graph matching solver to be differentiable and used feature hierarchies.
PCA~\cite{wang2019learning} differentiates end-to-end and learns linear and quadratic affinity costs.
QC-DGM~\cite{gao2021deep} proposes a differentiable quadratic constrained-optimization compatible with a deep learning optimizer and a balancing term in the loss function.
GLMNet~\cite{jiang2019glmnet} utilizes a GNN and alleviates oversmoothing by utilizing an anisotropic Laplacian ''sharpening'' operation.
CIE~\cite{yu2019learning} employs attention and improves upon plain attention by enforcing channel independence and sparsity in the ensuing matching decoding step.
DGMC~\cite{fey2020deep} uses the graph neural network~\cite{fey2018splinecnn} and an ad-hoc message passing routine for obtaining correspondences.
COMMON~\cite{lin2023graph} likewise uses SplineCNN and trains using contrastive losses.
ASAR~\cite{ren2022appearance} improves performance by using adversarial training and an advanced regularization technique.
GMTR~\cite{guo2024gmtr} uses a Transformer architecture with self- and cross-attention to exchange information between keypoints in the same and different images.

Crucially, while classical geometric verification (e.g., RANSAC) and specialized dense matchers~\cite{sarlin2020superglue,lindenberger2023lightglue,sun2021loftr} dominate rigid scene matching, our NMT is purpose-built for the extreme intra-class variation of sparse semantic matching. By jointly combining a Swin Transformer, SplineCNN structural priors, and hyperspherical normalized attention, NMT overcomes the limitations of pure attention models and achieves better performance while needing shorter training.

\begin{figure}[t]
    \begin{center}
        \includegraphics[width=0.79\linewidth]{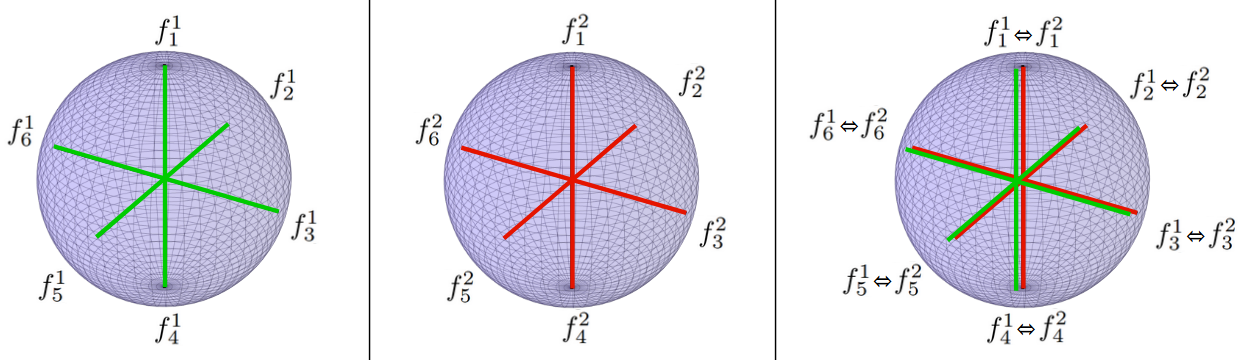}
    \end{center}
    \caption{Geometric illustration of hyperspherical and InfoNCE losses.
    The hyperspherical losses (two left spheres) from~\eqref{eq:prot_loss_s} distribute different keypoint features $f^i_j$ for different keypoints $j\in [m]$ within each image $i \in [2]$ across the hypersphere and are applied to each image separately.
    The InfoNCE (right side) loss from~\eqref{eq:infoNCE} aligns features $f^1_j \Leftrightarrow f^2_j$ from matching keypoints (assuming the matching is identity here) from different images.}
        \label{fig:losses-geometry}
\end{figure}

\section{Method}
\label{sec:method}

Our method consists of five building blocks:
\begin{description}
    \item[Visual feature extraction:]
    Visual features are extracted from the images using a pre-trained Swin Transformer~\cite{liu2021swin} backbone.
    \item[GNN:]
    Keypoints are treated as nodes to construct an undirected graph from the visual features.
    A SplineCNN~\cite{fey2018splinecnn} with shared weights is employed to refine the node features by aggregating local spatial information, processing each image independently.
    This approach was pioneered in~\cite{fey2020deep}.
    \item[Normalized Transformer:]
    The normalized Transformer (nGPT)~\cite{loshchilov2024ngpt} architecture is used with self-attention layers for exchanging information between keypoints of the same image and cross-attention layers for exchanging information between keypoints from different images.
    \item[Sinkhorn matching:]
    Using the refined features we compute cosine similarities between pairs of keypoints from each image.
    This similarity score matrix is passed through a log-space Sinkhorn algorithm for ensuring a doubly stochastic matrix.
    Each index of the maximum in each row indicates the best possible match between the source (row) and target (column) node.
    \item[InfoNCE and hyperspherical losses:]
    We use InfoNCE~\cite{oord2018representation} and hyperspherical~\cite{melekhov2019dgc} loss functions for shaping better feature representations.
    InfoNCE aligns features for the corresponding keypoints in different images and penalizes alignment of non-corresponding features.
    The hyperspherical loss distributes keypoint features from the same image uniformly on the hypersphere, ensuring more distinctive features.
    For better results we apply the hyperspherical loss after each normalized Transformer layer.
\end{description}

An illustration of our approach is provided in Figure~\ref{fig:matching-pipeline} for the inference and~\ref{fig:losses-pipeline} and~\ref{fig:losses-geometry} for losses during training.

\paragraph{Visual Feature Extraction.}
We are given two images $I^1, I^2$
alongside coordinates $k^1_1,\ldots,k^1_m$ and $k^2_1,\ldots,k^2_m$ specifying the $m$ keypoint positions in each image.
We pass both images through a Swin Transformer~\cite{liu2021swin} and obtain down-sampled features.
The spatial features corresponding to the specified keypoints are interpolated from these downsampled features using a bilinear sampling technique.
For each keypoint we extract features from the last and second last layer of the backbone and concatenate them.
Following BBGM~\cite{rolinek2020deep} we additionally mean-pool all features from the backbone to get a global feature for each image
that helps to class-condition the matching process.
We use the Swin-Large version as backbone.

\paragraph{GNN.}
To incorporate the spatial structure of the objects as indicated by the geometry of the keypoints, we use a SplineCNN~\cite{fey2018splinecnn} as suggested in~\cite{fey2020deep}.
We construct a graph with nodes being keypoints and edges coming from a Delaunay triangulation.
Two rounds of graph convolution are performed with shared weights to refine feature representations for each image independently.
Unlike general-purpose GNNs that focus on aggregating features across nodes without explicit consideration of their spatial layout, SplineCNN uses spline-based kernels that adapt to the graph's Euclidean geometry, allowing for a better representation of spatial interactions.

The GNN operates on a message-passing paradigm, where node features are iteratively updated by aggregating information from neighboring nodes at each layer.
Trainable B-Spline kernels anisotropically convolve features from other nodes.

Our implementation employs two spline convolution layers, each with a kernel size of 5 and uses max aggregation.
We use ReLU as non-linearity.
We use Euclidean coordinates for the geometric input to the trainable B-splines.

\paragraph{Normalized Transformer.}
The normalized Transformer~\cite{loshchilov2024ngpt}, a variant of the original Transformer architecture~\cite{vaswani2017attention}, uses hyperspherical normalization and projects intermediate and final representations back to unit norm.
This aligns well with the graph matching setting where we want to measure potential correspondences by cosine similarity of their (implicitly normalized) features.
Experiments in NLP have demonstrated that the normalized Transformer architecture can converge faster, be numerically more stable, and reach better solutions.
Constraining embeddings to a unit hypersphere mitigates the ``hubness'' problem common in high-dimensional spaces, forcing the network to optimize angular margins directly rather than relying on unbounded feature magnitudes.

In particular, the normalized Transformer uses normalization after attention, the MLP block and the residual connections.
Let $f = f_1,\ldots,f_m$ be a feature sequence coming from all the keypoints in an image.
For cross-attention let $f_{o}$ be the keypoint feature sequence from the other image.
Then normalized self-attention, cross-attention and MLP layers can be formulated as follows, where element-wise step sizes $\alpha_A, \alpha_C, \alpha_M$ in residual connections are learned positive vectors:

\begin{align}
    f_A &\leftarrow \text{Norm}(\text{Self-Attn}(f)), \quad f \leftarrow \text{Norm}(f + \alpha_A \cdot (f_A - f)) \label{eq:self-attn} \\
    f_A &\leftarrow \text{Norm}(\text{Cross-Attn}(f, f_o)), \quad f \leftarrow \text{Norm}(f + \alpha_C \cdot (f_A - f)) \label{eq:cross-attn} \\
    f_M &\leftarrow \text{Norm}(\text{MLP}(f)), \quad f \leftarrow \text{Norm}(f + \alpha_M \cdot (f_M - f)) \label{eq:mlp}
\end{align}

A Transformer layer in our matching method consists of
first doing normalized self-attention separately for keypoint features of each image,
then doing two normalized cross-attention passes where one keypoint feature sequence attends to the other, and finally passes through a normalized MLP block.
Additionally, after cross-attention we element-wise modulate keypoint features with the global feature token and normalize afterwards to unit norm again.

We use 4 such two-stream normalized Transformer decoder layers using 12 heads with a feature dimension of $648$. We use SiLU activations.

\paragraph{Matching.}
To establish correspondences between keypoints, we first compute cosine similarities between each pair of features coming from different images.
During inference this square affinity matrix is given to the Sinkhorn algorithm, which outputs a doubly stochastic matrix.
To decode final correspondences, we use the computed doubly stochastic matrix and go through each row $i$ and pick the column $j$ with maximum entry, meaning that keypoint $i$ in the first image is matched to keypoint $j$ in the second one.

Our full matching pipeline is detailed in the Normalized Matching Transformer algorithm.

\begin{algorithm}[!ht]
    \caption{\texttt{Normalized Matching Transformer}}
    \label{alg:normalized-matching-transformer}
    \KwInput{Input images $I^1,I^2$,\\ keypoints $k_1^1,\ldots,k^1_m$, $k^2_1,\ldots,k^2_m$}
    \KwOutput{Matching $\pi: [m] \rightarrow [m]$}
    \tcp{Swin Transformer backbone}
    $g^i_1,\ldots,g^i_n = \text{Backbone}(I^i)$ \hhfill $\forall i\in[2]$\;
    \tcp{Global feature token}
    $f^i_{global} = \text{Avg-Pool}(g^i_1,\ldots,g^i_n)$ \hhfill $\forall i\in[2]$\;
    \tcp{Interpolate features at keypoint}
    $f^i_j = \text{Interp}(g^i,k_j)$, \hhfill $\forall i\in[2]$, $j\in[m]$\;
    \tcp{SplineCNN GNN (shared weights)}
    $f^i = f^i_1,\ldots,f^i_m = \text{GNN}(f^i_1,\ldots,f^i_m)$ \hhfill $\forall i\in[2]$\;
    \tcp{Normalized Transformer Decoder}
    \For{$iter=1,\ldots,L$}{
        $f^i = \texttt{Norm.\,Self-Attn}(f^i,f^i_{global})$\;
        $f^1 = \texttt{Norm.\,Cross-Attn}(f^1,f^2)$\;
        $f^2 = \texttt{Norm.\,Cross-Attn}(f^2,f^1)$\;
        $f^i_j = \text{Norm}(f^i_j \cdot f^i_{global})$ \hhfill $i\in[2],j\in[m]$ \;
        $f^i,f^i_{global} = \texttt{Norm.\,MLP}(f^i, f^i_{global})$ \hhfill $\forall i\in [2]$\;}

    \tcp{Sinkhorn matching}
    $C_{ij} = \cossim(f^1_i, f^2_j)$ \hhfill  $\forall i,j \in [m]$\;
    $A = \operatorname{Sinkhorn}(C)$\;
    $\pi(i) = \operatorname{arg\,max}_{j \in [m]} \{ A_{ij} \}$ \hhfill  $\forall i \in [m]$\;
\end{algorithm}

\paragraph{Training Losses.}

We use the contrastive InfoNCE loss introduced in~\cite{oord2018representation} for better feature representations.
Corresponding points are treated as positive pairs, while non-corresponding matches from the other image are treated as negative ones.
This leads to matching keypoints having aligned representations with large cosine similarity while non-matching ones having low cosine similarity.

In particular, let $f^1_i$ and $f^2_j$ be two matching keypoint features coming out of the Transformer decoder.
Let $f^1_i$ and $f^2_{l}, l \in [m] \backslash \{j\}$ be the keypoint features in image $2$ that do not match to $f^1_i$.
Then the InfoNCE loss is
\begin{equation}
    \mathcal{L}_{\text{InfoNCE}} = -\log \frac{\exp(\cossim(f^1_i, f^2_j) / \tau)}{\sum_{l \in [m] \backslash \{j\}} \exp(\cossim(f^1_i, f^2_l) / \tau)},
    \label{eq:infoNCE}
\end{equation}
Here $\tau > 0$ is a learnable parameter.
The overall InfoNCE loss is then the summation of the InfoNCE losses over all keypoints features.
To symmetrize, we also take the matches the other way around.

To further encourage separation between keypoint features coming from the same image and to promote a more uniform distribution of features on the hypersphere we use a hyperspherical loss~\cite{mettes2019hyperspherical}.
Let
\begin{equation}
   S^i = \left(\cossim(f^i_j, f^i_k) \right)_{j,k \in [m],\, j \neq k} \in \mathbb{R}^{m \times (m-1)}
\end{equation}
be the matrix of all pairwise cosine similarities between different keypoints within image $i$.
Then the hyperspherical loss for image $i$ is
\begin{equation}
    \mathcal{L}_{\mathbf{HS}}^i = \sum_{j=1}^m \max_{k \neq j} S^i \,.
    \label{eq:prot_loss_s}
\end{equation}
This loss penalizes whenever two different keypoints within the same image are aligned. The $\max_{k \neq j}$ excludes the self-similarity term, ensuring a keypoint is not penalized for aligning with itself.
The overall hyperspherical loss averages over both images: $\mathcal{L}_{\mathbf{HS}} = \frac{1}{2}\sum_{i=1}^{2} \mathcal{L}_{\mathbf{HS}}^i$.

We also incorporate the hyperspherical loss as an auxiliary layer loss on every Transformer layer. The loss is weighted by a factor that decreases linearly with layer depth: layer $k$ receives weight $(L - k + 1) \cdot p$, where $p = 0.3$ is the step size. For $L=4$ layers this yields weights $1.2, 0.9, 0.6, 0.3$ from the first to the last layer, applying stronger regularization to shallower layers where features are less refined. Then the loss is
\begin{equation}
    \mathcal{L}_{\mathbf{HS}}^{\mathbf{layer}} = \sum_{k=1}^L (L - k + 1) \cdot p \cdot \mathcal{L}_{\mathbf{HS}}^{(k)}\,,
    \label{eq:prot_loss_l}
\end{equation}
where $\mathcal{L}_{\mathbf{HS}}^{(k)}$ is the layer-wise hyperspherical loss at layer $k$.
We then average over the two decoders to obtain a single layer-wise loss and add that to the overall hyperspherical loss computed from the final decoder outputs. Consequently, this strategy encourages a more uniform distribution of keypoint features on the hypersphere across all layers, ultimately leading to enhanced feature distinctiveness and improved matching performance.

Our overall loss sums up the InfoNCE and hyperspherical loss without any weighting.

An illustration of the loss construction is given in Figures~\ref{fig:losses-pipeline} and~\ref{fig:losses-geometry}.

\section{Experiments}
\paragraph{Training Details and Design Choices.}

Our network is trained using the Adam optimizer~\cite{kingma2014adam}. The initial learning rate for the network is set to \(5 \times 10^{-4}\), while the Swin Transformer~\cite{liu2021swin} backbone with layer normalization uses a learning rate scaled down by a factor of \(0.03\). Additional normalization was omitted due to the inherent design of the normalized Transformer. The learning rate is scaled by a factor of \(0.1\) after epoch 2 and 5. For PascalVOC the batch size of image pairs is set to 8 and for SPair-71k the batch size is set to 5. Our validation set is obtained by taking $1{,}000$ image pairs per class. We use augmentations provided through the Albumentations package~\cite{info11020125}. Specifically, we use Mixup~\cite{zhang2017mixup}, Cutmix~\cite{yun2019cutmix} and Random Erasing~\cite{zhong2020random}.

To determine the optimal architecture and training hyperparameters, we initially conducted a coarse grid search to identify a high-performing configuration scope, followed by manual empirical fine-tuning to balance accuracy with hardware limitations. To accommodate our GPU memory constraints (40 GB A100 limit) while maximizing performance, we configured the Swin-Large backbone with a patch size of 4, window size of 24, and embedding dimension of 128. The SplineCNN employs 2 layers with a kernel size of 5, which we empirically found sufficiently captures local keypoint geometries without over-smoothing. The Normalized Transformer uses 4 decoder layers, 12 attention heads, and a feature dimension of $d_{\text{model}} = 648$ with SiLU activations. The model is trained on one A100 GPU and takes about 9 hours for PascalVOC and 7 hours for SPair-71k.

Notably, our model requires only 6 epochs to converge, whereas BBGM~\cite{rolinek2020deep} requires 10, ASAR~\cite{ren2022appearance} requires 16 and COMMON~\cite{lin2023graph} requires 16 epochs.
Even though time per epoch may not be directly comparable, since the normalized Transformer requires somewhat more time due to suboptimal kernel fusion operations compared to a vanilla Transformer, this reveals significant potential training time savings on a more optimized normalized Transformer implementation.

\begin{table*}[!t]
\caption{Average accuracy (\%) of each object category on Pascal VOC. Best results are \textbf{bold} and second best are \underline{underlined}.}
    \centering
    \resizebox{\textwidth}{!}{
    \begin{tabular}{rccccccccccccccccccccc}
        \toprule
        \textbf{Method} & \includegraphics[height=8pt]
        {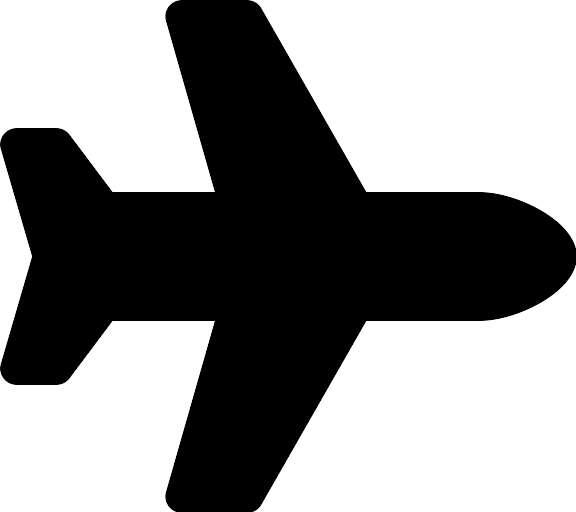} & \includegraphics[height=8pt]{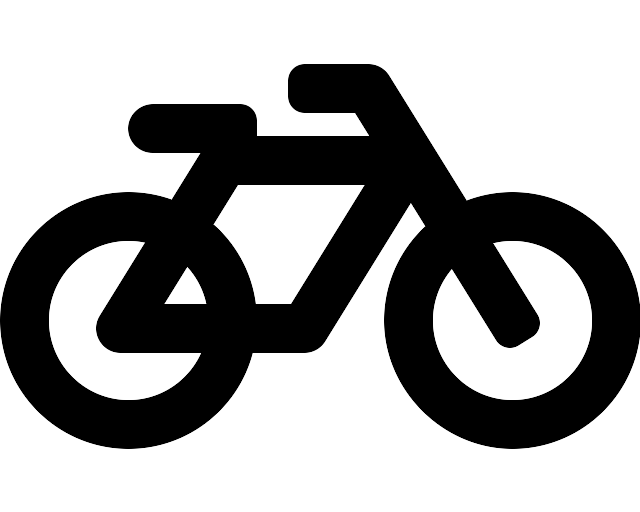} & \includegraphics[height=8pt]{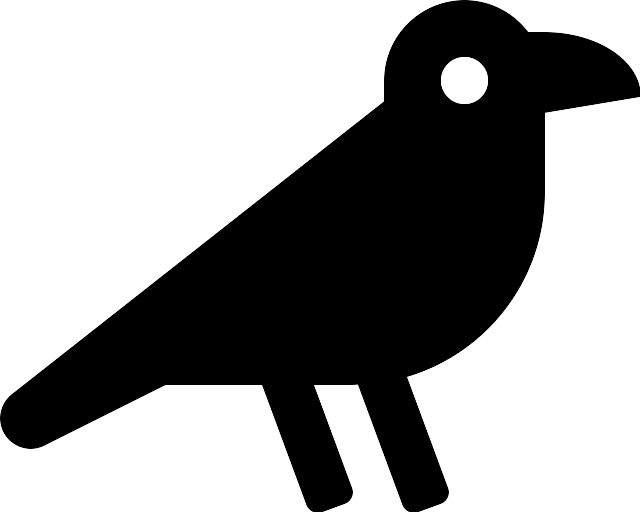} & \includegraphics[height=8pt]
        {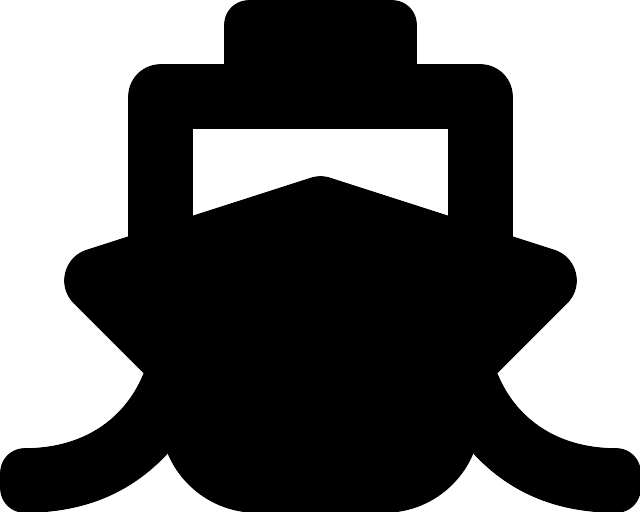} & \includegraphics[height=8pt]{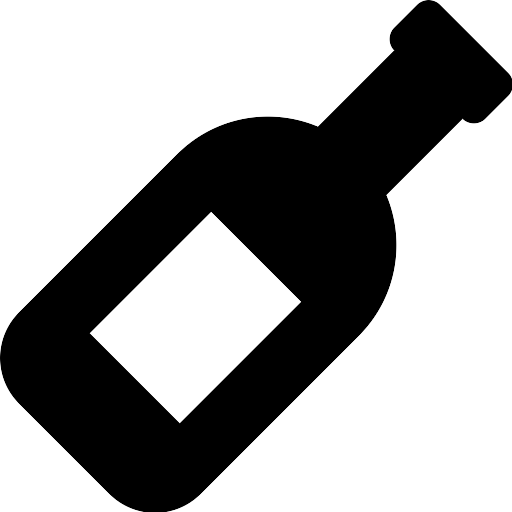} & \includegraphics[height=8pt]{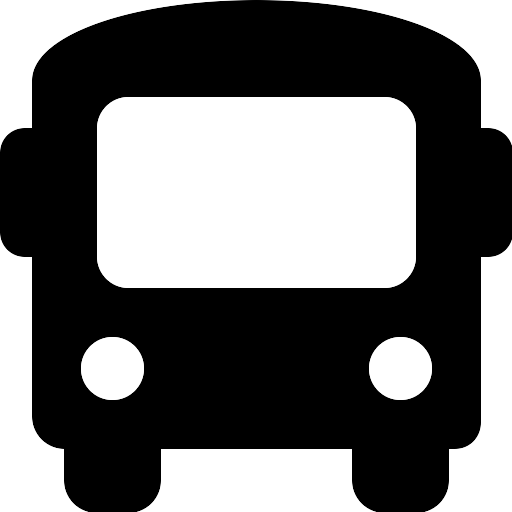} & \includegraphics[height=8pt]{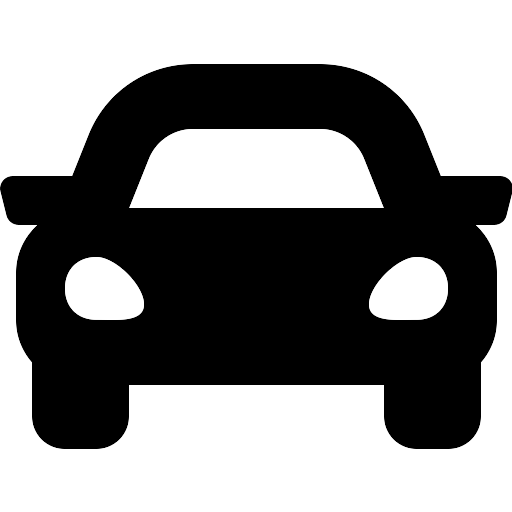} & \includegraphics[height=8pt]{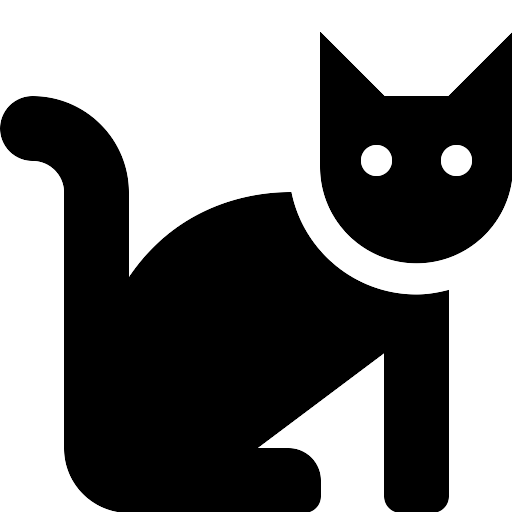} & \includegraphics[height=8pt]{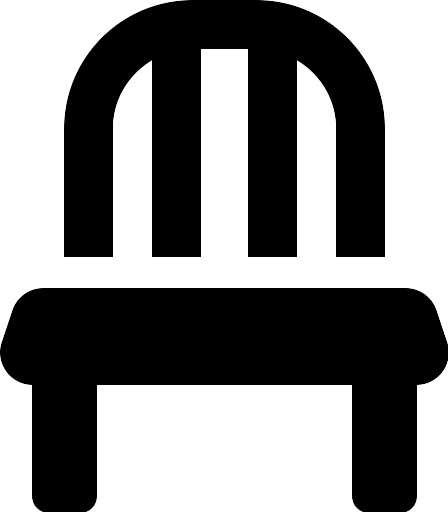} & \includegraphics[height=8pt]{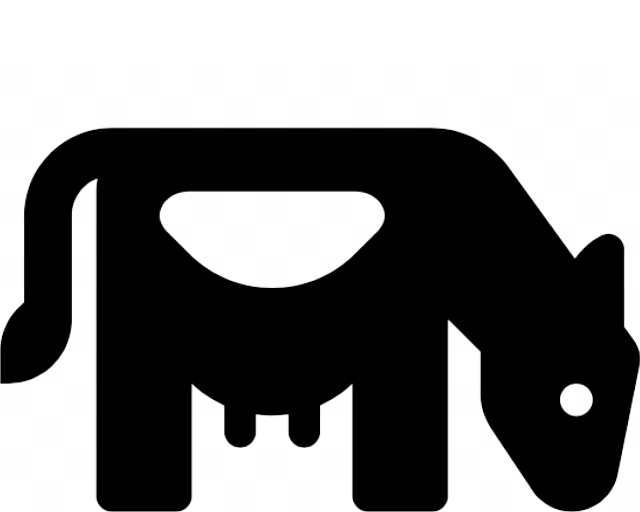} & \includegraphics[height=8pt]{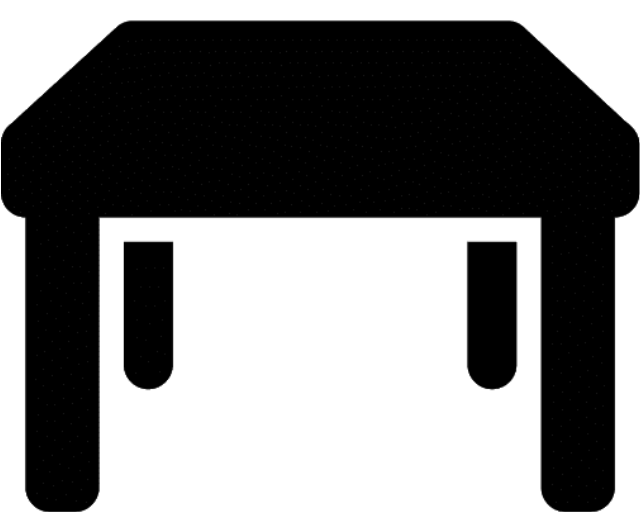} & \includegraphics[height=8pt]{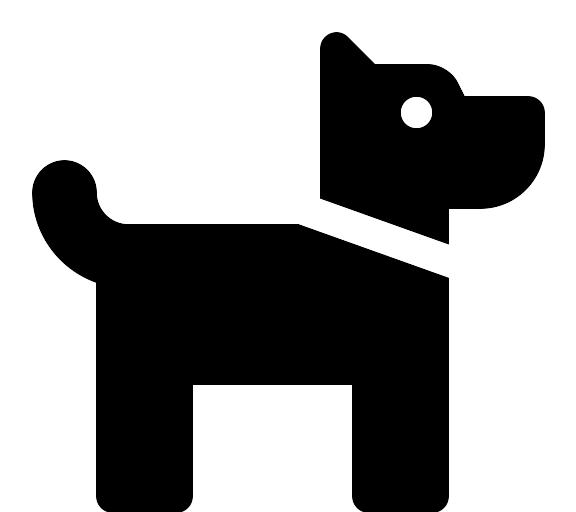} & \includegraphics[height=8pt]{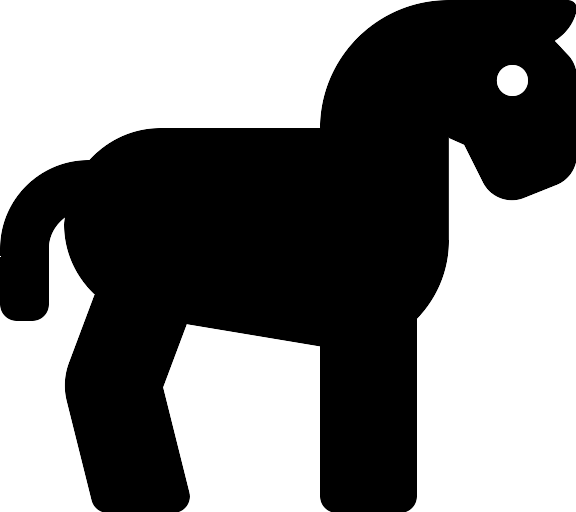} & \includegraphics[height=8pt]{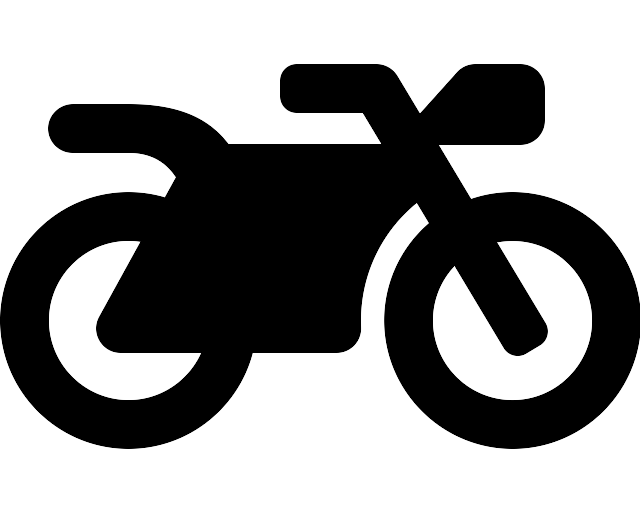} & \includegraphics[height=8pt]{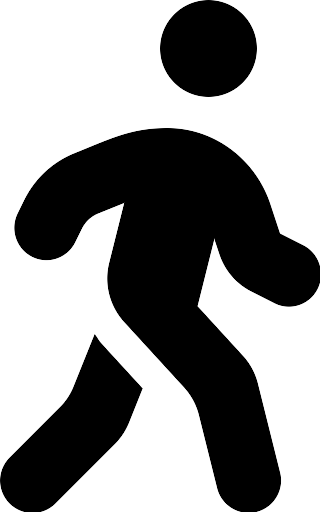} & \includegraphics[height=8pt]{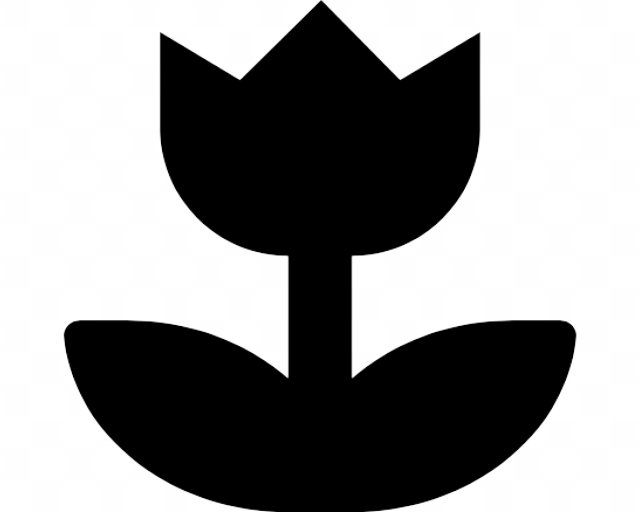} & \includegraphics[height=8pt]{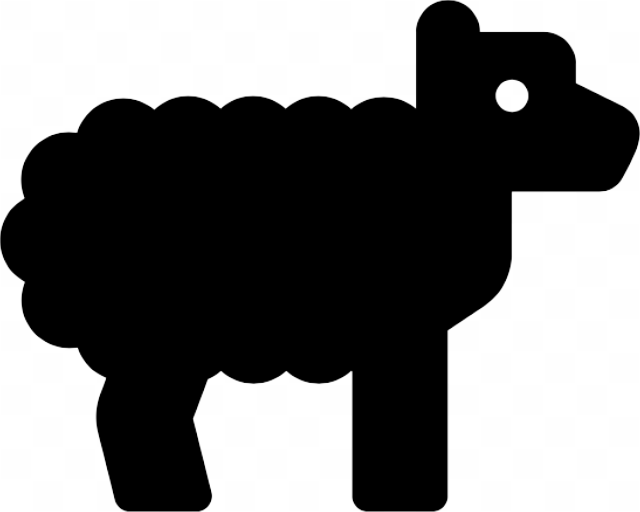} & \includegraphics[height=8pt]
        {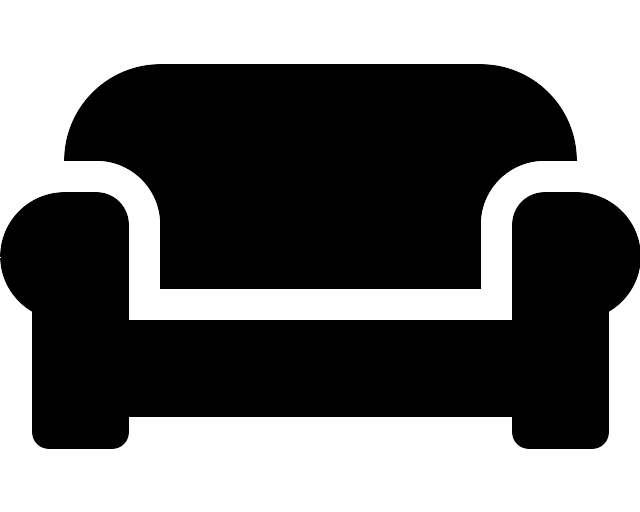} & \includegraphics[height=8pt]{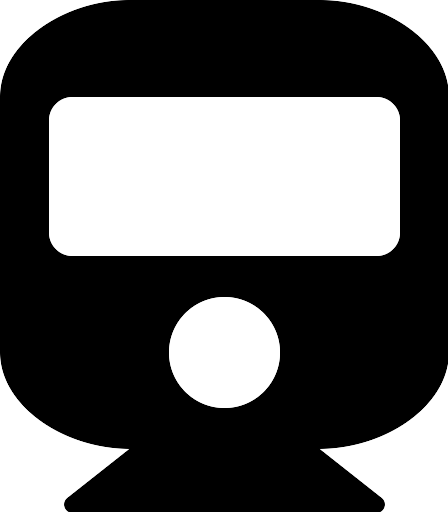} & \includegraphics[height=8pt]{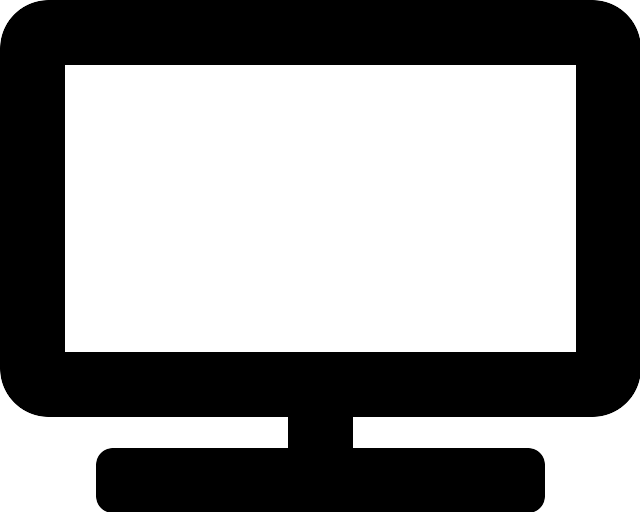} & \textbf{Mean} \\
        \midrule
        GMN-PL~\cite{patil2021layoutgmn} & 31.1 & 46.2 & 58.2 & 45.9 & 70.6 & 76.5 & 61.2 & 61.7 & 35.5 & 53.7 & 58.9 & 57.5 & 56.9 & 49.3 & 34.1 & 77.5 & 57.1 & 53.6 & 83.2 & 88.6 & 57.9\\
        PCA~\cite{wang2019learning} & 40.9 & 55.0 & 65.8 & 47.9 & 76.9 & 77.9 & 63.5 & 67.4 & 33.7 & 66.5 & 63.6 & 61.3 & 58.9 & 62.8 & 44.9 & 77.5 & 67.4 & 57.5 & 86.7 & 90.9 & 63.8\\
        NGM~\cite{wang2021neural} & 50.8 & 64.5 & 59.5 & 57.6 & 79.4 & 76.9 & 74.4 & 69.9 & 41.5 & 62.3 & 68.5 & 62.2 & 62.4 & 64.7 & 47.8 & 78.7 & 66.0 & 63.3 & 81.4 & 89.6 & 66.1\\
        GLMNet~\cite{jiang2019glmnet} & 52.0 & 67.3 & 63.2 & 57.4 & 80.3 & 74.6 & 70.0 & 72.6 & 38.9 & 66.3 & 77.3 & 65.7 & 67.9 & 64.2 & 44.8 & 86.3 & 69.0 & 61.9 & 79.3 & 91.3 & 67.5\\
        CIE~\cite{yu2019learning} & 51.2 & 69.2 & 70.1 & 55.0 & 82.8 & 72.8 & 69.0 & 74.2 & 39.6 & 68.8 & 71.8 & 70.0 & 71.8 & 66.8 & 44.8 & 85.2 & 69.9 & 65.4 & 85.2 & 92.4 & 68.9\\
        DGMC~\cite{fey2020deep} & 50.4 & 67.6 & 70.7 & 70.5 & 87.2 & 85.2 & 82.5 & 74.3 & 46.2 & 69.4 & 69.9 & 73.9 & 73.8 & 65.4 & 51.6 & 98.0 & 73.2 & 69.6 & 94.3 & 89.6 & 73.2$\pm$0.5\\
        BBGM~\cite{rolinek2020deep} & 61.5 & 75.0 & 78.1 & \underline{80.0} & 87.4 & 93.0 & 89.1 & 80.2 & 58.1 & 77.6 & 76.5 & 79.3 & 78.6 & 78.8 & 66.7 & 97.4 & 76.4 & 77.5 & 97.7 & \underline{94.4} & 80.1$\pm$0.6\\
        GMTR~\cite{guo2024gmtr} & \underline{69.0} & 74.2 & \underline{84.1} & 75.9 & 87.7 & \underline{94.2} & \underline{90.9} & \underline{87.8} & \underline{62.7} & \underline{83.5} & \underline{93.9} & \underline{84.0} & 78.7 & \underline{79.6} & \underline{69.2} & \textbf{99.3} & \underline{82.5} & \underline{83.0} & \underline{99.1} & 93.3 & \underline{83.6} \\
        COMMON~\cite{lin2023graph} & 65.6 & \underline{75.2} & 80.8 & 79.5 & \underline{89.3} & 92.3 & 90.1 & 81.8 & 61.6 & 80.7 & \textbf{95.0} & 82.0 & \underline{81.6} & 79.5 & 66.6 & \underline{98.9} & 78.9 & 80.9 & \textbf{99.3} & 93.8 & 82.7 \\

        \bottomrule
        \addlinespace[2pt]
        \textbf{NMT} (ours) & \textbf{75.8} & \textbf{81.9} & \textbf{90.9} & \textbf{82.4} & \textbf{93.5} & \textbf{95.4} & \textbf{92.7} & \textbf{90.7} & \textbf{84.6} & \textbf{85.2} & 92.9 & \textbf{89.3} & \textbf{89.4} & \textbf{86.9} & \textbf{77.2} & 98.5 & \textbf{85.8} & \textbf{88.0} & 97.6 & \textbf{95.5} & \textbf{88.7} \\

    \end{tabular}
    }
    \label{tab:performance_pascalVOC}
\end{table*}

\begin{table*}[!t]
\caption{Average accuracy (\%) of each object category on SPair-71k. Best results are \textbf{bold} and second best are \underline{underlined}.}
\scriptsize
    \resizebox{\textwidth}{!}{
    \begin{tabular}{rccccccccccccccccccc}
        \toprule
        \textbf{Method} & \includegraphics[height=8pt]
        {./graphics/class_icons/aeroplane.png} & \includegraphics[height=8pt]{./graphics/class_icons/bicycle.png} & \includegraphics[height=8pt]{./graphics/class_icons/bird.png} & \includegraphics[height=8pt]
        {./graphics/class_icons/boat.png} & \includegraphics[height=8pt]{./graphics/class_icons/bottle.png} & \includegraphics[height=8pt]{./graphics/class_icons/bus.png} & \includegraphics[height=8pt]{./graphics/class_icons/car.png} & \includegraphics[height=8pt]{./graphics/class_icons/cat.png} & \includegraphics[height=8pt]{./graphics/class_icons/chair.png} & \includegraphics[height=8pt]{./graphics/class_icons/cow.png} & \includegraphics[height=8pt]{./graphics/class_icons/dog.png} & \includegraphics[height=8pt]{./graphics/class_icons/horse.png} & \includegraphics[height=8pt]{./graphics/class_icons/motorbike.png} & \includegraphics[height=8pt]{./graphics/class_icons/person.png} & \includegraphics[height=8pt]{./graphics/class_icons/pottedplant.png} & \includegraphics[height=8pt]{./graphics/class_icons/sheep.png} & \includegraphics[height=8pt]{./graphics/class_icons/train.png} & \includegraphics[height=8pt]{./graphics/class_icons/tvmonitor.png} & \textbf{Mean} \\
        \midrule
        DGMC~\cite{fey2020deep} & 54.8 & 44.8 & 80.3 & 70.9 & 65.5 & 90.1 & 78.5 & 66.7 & 66.4 & 73.2 & 66.2 & 66.5 & 65.7 & 59.1 & 98.7 & 68.5 & 84.9 & 98.0 & 72.2\\
        BBGM~\cite{rolinek2020deep} & 66.9 & 57.7 & 85.8 & 78.5 & 66.9 & 95.4 & 86.1 & 74.6 & 68.3 & 78.9 & 73.0 & 67.5 & 79.3 & 73.0 & 99.1 & 74.8 & 95.0 & 98.6 & 78.9\\
        GMTR~\cite{guo2024gmtr} & 75.6 & 67.2 & \underline{92.4} & 76.9 & 69.4 & 94.8 & 89.4 & 77.5 & 72.1 & \underline{86.3} & 77.5 & 72.2 & \textbf{86.4} & \underline{79.5} & \underline{99.6} & \textbf{84.4} & 96.6 & 99.7 & 83.2\\
        COMMON~\cite{lin2023graph} & \underline{77.3} & \underline{68.2} & 92.0 & \underline{79.5} & \underline{70.4} & \underline{97.5} & \underline{91.6} & \textbf{82.5} & \underline{72.2} & \textbf{88.0} & \underline{80.0} & \underline{74.1} & \underline{83.4} & \textbf{82.8} & \textbf{99.9} & \textbf{84.4} & \underline{98.2} & \underline{99.8} & \underline{84.5}\\
        \bottomrule
        \addlinespace[2pt]

        \textbf{NMT} (ours) & \textbf{79.3} & \textbf{72.7} & \textbf{94.9} & \textbf{84.2} & \textbf{74.8} & \textbf{98.7} & \textbf{94.4} & \underline{82.2} & \textbf{81.1} & \textbf{88.0} & \textbf{85.5} & \textbf{81.3} & 82.3 & 79.4 & \underline{100} & \underline{83.2} & \textbf{99.2} & \textbf{99.9} & \textbf{86.7}\\

    \end{tabular}
    }
    \label{tab:performance_SPair}
\end{table*}

\paragraph{Datasets.}
We train and test on PascalVOC and SPair-71k, the two most challenging current sparse semantic keypoint matching datasets.
We opted to forego e.g.\ Willow Object Class~\cite{cho2013learning} and other similar datasets since performance of previous works is already almost perfect there.

\begin{description}
    \item[PascalVOC:]
    We use PascalVOC~\cite{everingham2010pascal} images with Berkeley annotations~\cite{bourdev2009poselets}.
    Images are from 20 classes and are of size $256 \times 256$.
    Up to 23 keypoints are contained in each image.
    In order to be comparable to other works we use standard intersection filtering, i.e.\ when matching we only include keypoints that are in both images and discard outliers.
    \item[SPair-71k:]
    The SPair-71k dataset~\cite{min2019spair} is a successor to PascalVOC and offers higher image quality and keypoint annotation and removal of problematic and poorly annotated image categories.
    It contains $70{,}958$ image pairs.
    Images are taken from PascalVOC and Pascal3D+.
\end{description}

\paragraph{Baselines.}
We compare our results with the highest-performing baselines from the literature, to the best of our knowledge.

\begin{figure}[!htbp]
    \centering
    % --- First Row (Perfect matchings) ---
    \begin{subfigure}[b]{0.3\textwidth}
        \centering
        \includegraphics[height=3cm, width=\textwidth, keepaspectratio]{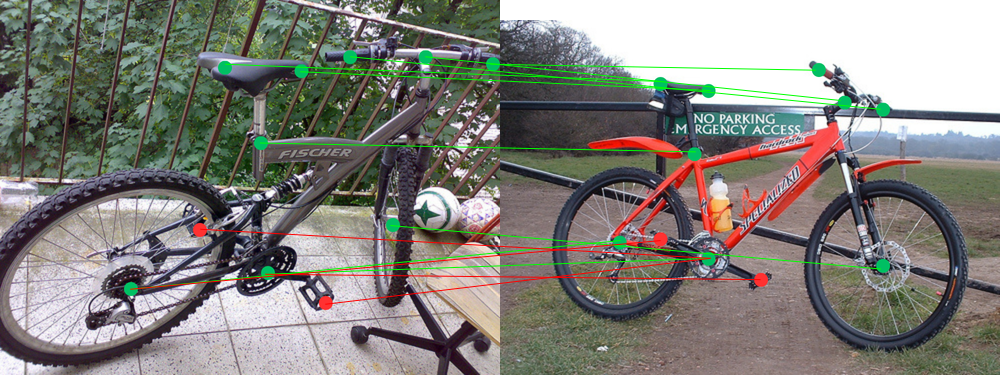}
        \caption{Bikes}
    \end{subfigure}
    \hfill
    \begin{subfigure}[b]{0.3\textwidth}
        \centering
        \includegraphics[height=3cm, width=\textwidth, keepaspectratio]{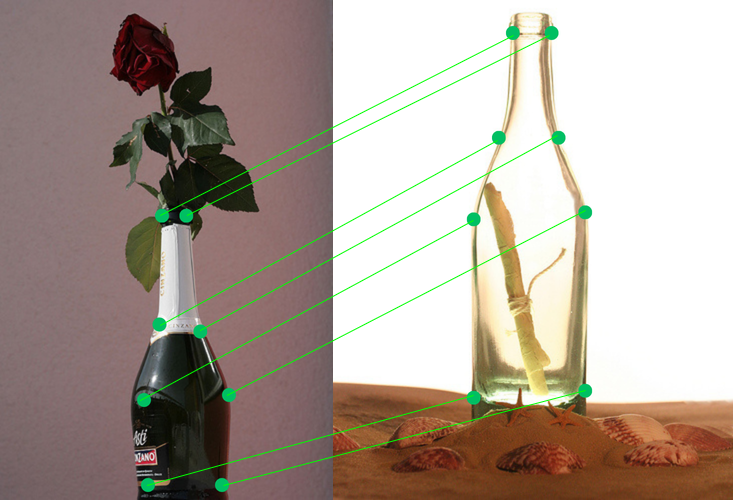}
        \caption{Bottles}
    \end{subfigure}
    \hfill
    \begin{subfigure}[b]{0.3\textwidth}
        \centering
        \includegraphics[height=3cm, width=\textwidth, keepaspectratio]{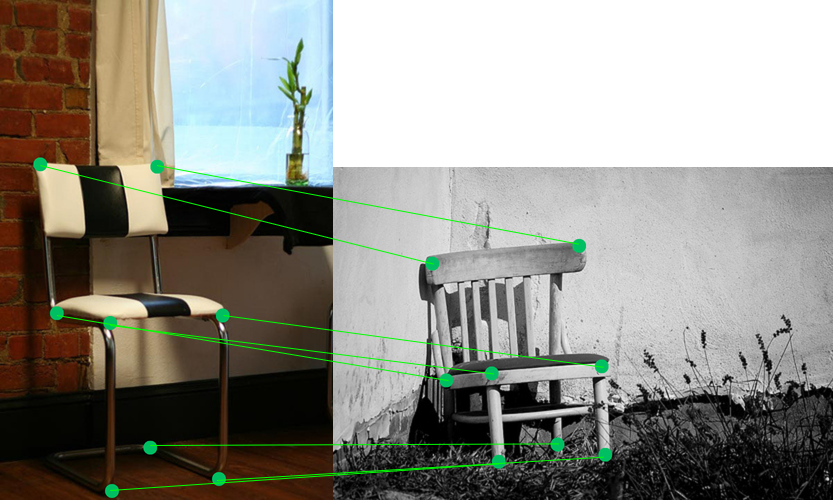}
        \caption{Chairs}
    \end{subfigure}

    \vspace{1em}

    % --- Second Row (Failure Cases) ---
    \begin{subfigure}[b]{0.3\textwidth}
        \centering
        \includegraphics[height=3cm, width=\textwidth, keepaspectratio]{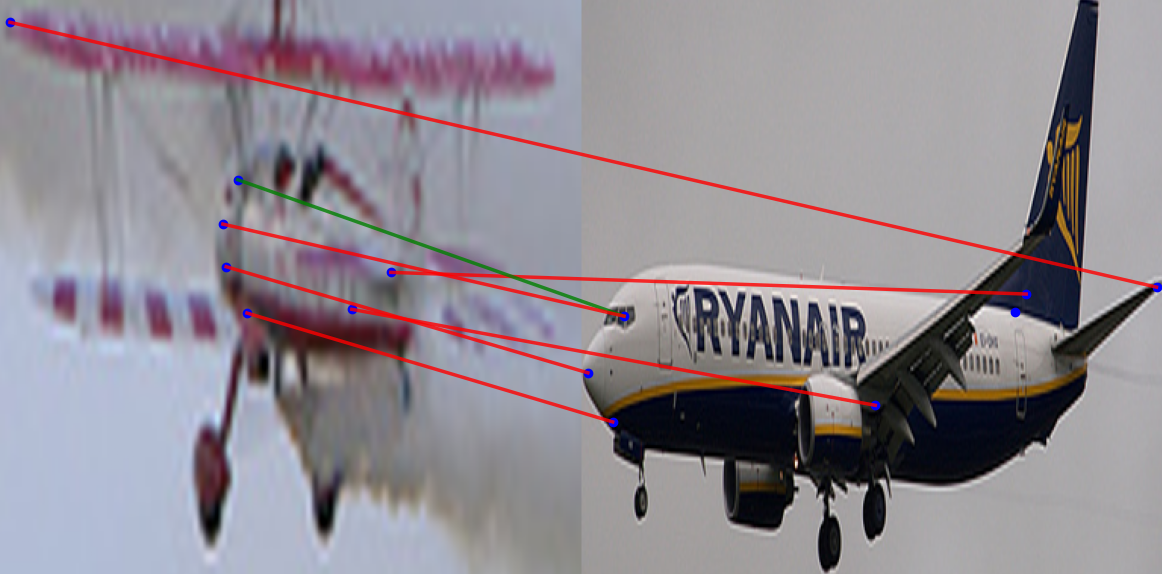}
        \caption{Aeroplane (Failure)}
    \end{subfigure}
    \hfill
    \begin{subfigure}[b]{0.3\textwidth}
        \centering
        \includegraphics[height=3cm, width=\textwidth, keepaspectratio]{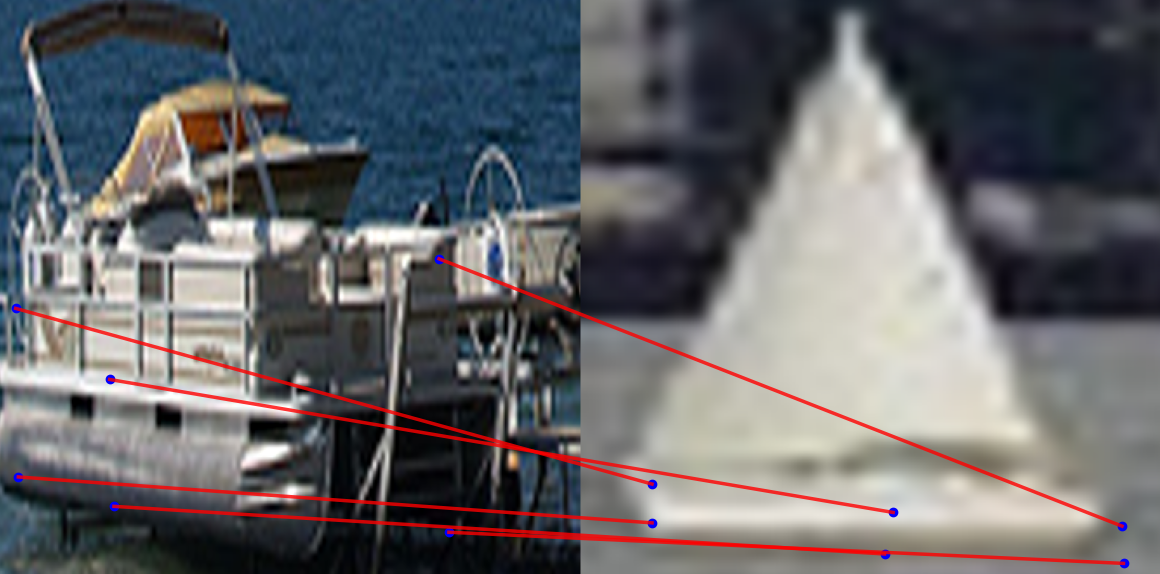}
        \caption{Boat (Failure)}
    \end{subfigure}
    \hfill
    \begin{subfigure}[b]{0.3\textwidth}
        \centering
        \includegraphics[height=3cm, width=\textwidth, keepaspectratio]{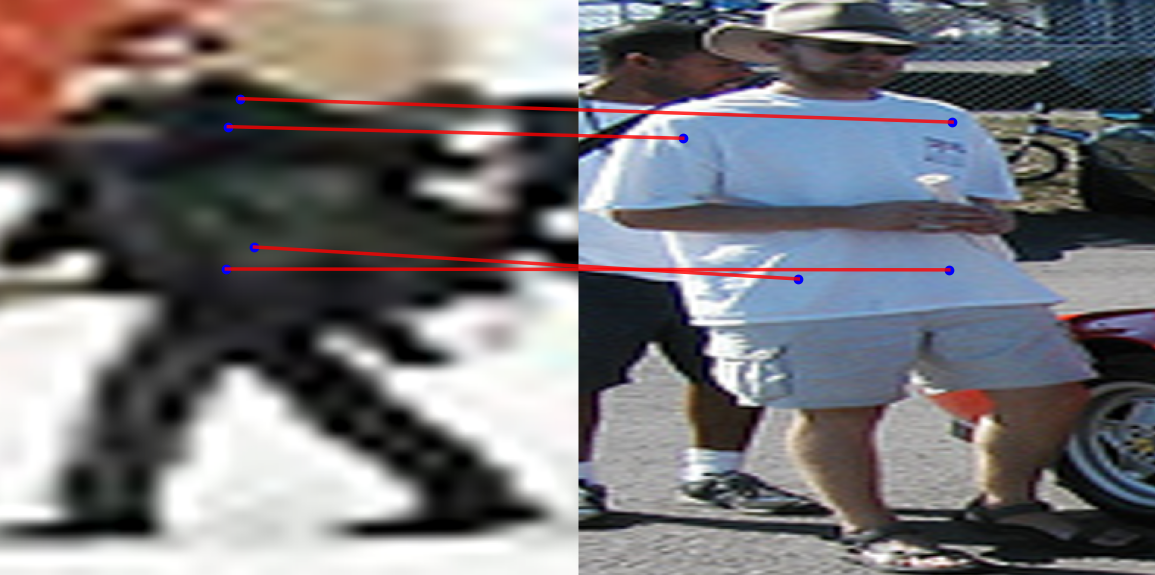}
        \caption{Person (Failure)}
    \end{subfigure}

    \caption{Qualitative results of selected keypoint matchings from the SPair-71k~\cite{min2019spair} dataset. The top row depicts perfect matchings, while the bottom row showcases identified failure cases driven by severe image quality degradation.}
    \label{fig:qualitative-examples}
\end{figure}

\paragraph{Results.}
Class-specific and overall results in terms of matching accuracy are presented in Table~\ref{tab:performance_pascalVOC} for PascalVOC and in Table~\ref{tab:performance_SPair} for SPair-71k.
Selected qualitative examples are shown in Figure~\ref{fig:qualitative-examples}.

On PascalVOC we outperform on average the baselines by $5.1\%$.
We are better on 17 out of 20 image categories.
On SPair-71k we overall outperform all baselines by $2.2\%$ matching accuracy.
We are better on 13 out of 18 image categories and second best on 3.

\paragraph{Inference Speed.}
We measure inference speed on a single NVIDIA GeForce RTX 4090 (batch size = 1), averaging over 1000 forward passes.
The complete Normalized Matching Transformer processes each image pair in 44.4 ms, with the backbone + SplineCNN accounting for 39 ms and the two decoders for 5.4 ms.

\paragraph{Ablations.}
We provide ablations on our main architectural and training contributions on PascalVOC.
The results are summarised in Table~\ref{tab:ablation-pascal-voc}.
We see that all contributions significantly add to final performance.
The main performance driver is our losses, followed by the improved backbone, normalized Transformer and augmentations.
Augmentations, while giving $1.2\%$, are still significant, but not overly so.
The backbone ablation shows that we obtain better results on PascalVOC even when we employ the VGG backbone (83.8\%), also used by COMMON~\cite{lin2023graph}, ASAR~\cite{ren2022appearance} and BBGM~\cite{rolinek2020deep}, and we even slightly outperform GMTR~\cite{guo2024gmtr} which uses a Swin Transformer backbone.

We have also experimented with other simple pixel-wise augmentations like changing saturation value, random gamma, RGB shift etc., which however degraded performance.

\begin{table}[!t]
\caption{Ablation study on PascalVOC.
    We ablate augmentations, replacing the normalized Transformer by a vanilla one, replacing InfoNCE and hyperspherical loss by cross-entropy and replacing the Swin Transformer backbone with VGG~\cite{simonyan2014very}.
    }
    \centering
    \begin{tabular}{l c}
        \toprule
        \textbf{Method} & \textbf{PascalVOC} \\
        \midrule
        NMT (FULL)                          & 88.7\% \\
        \quad w/o augmentation              & -1.2\% \\
        \quad w/o layer loss                & -0.8\% \\
        \midrule
        \quad w/ vanilla Transformer        & -2.6\% \\
        \quad w/ cross-entropy loss         & -15.1\% \\
        \quad w/ VGG16                      & -4.9\% \\
        \bottomrule
    \end{tabular}
    \label{tab:ablation-pascal-voc}
\end{table}

\paragraph{Robustness to Localization Noise.}
Unlike rigid geometric matching, semantic matching must handle extreme appearance variations without relying on perfectly repeatable low-level textures. To evaluate NMT's zero-shot robustness to the localization inaccuracies typical of automated keypoint detectors (e.g., SIFT or SuperPoint), we injected artificial Gaussian jitter ($\sigma \in \{2, 5, 10\}$ pixels) into the ground-truth coordinates during inference on PascalVOC, resulting in accuracy drops of only 0.12\%, 0.50\%, and 1.41\%, respectively. This demonstrates that NMT effectively learns robust semantic regions and structural priors rather than relying strictly on exact coordinate precision.

\paragraph{Failure Modes.}
To provide a detailed qualitative analysis, we programmatically isolated the lowest-performing image pairs within the five most challenging classes (aeroplane, bicycle, boat, chair, and person). As shown in the bottom row of Figure~\ref{fig:qualitative-examples}, our jitter ablation confirms the model handles coordinate noise robustly. The primary failure driver is instead severe image quality degradation: when a high-resolution source image is paired with an exceptionally low-resolution or heavily artifacted target image, the visual feature drift becomes a bottleneck. If this severe quality imbalance is coupled with a minor viewpoint change, matching performance drops further. Resolving these extreme cross-resolution domain gaps remains an area for future work.

\paragraph{Exploratory Analysis on Dense Geometric Matching.}
While NMT is fundamentally designed for sparse semantic correspondences where object structure and appearance change drastically across instances, we conducted an exploratory evaluation to test its capabilities on dense, rigid geometric tasks. To this end, we generated a synthetic geometric dataset based on PascalVOC, applying homography magnitudes between 25\% and 75\% during training to mimic the extreme viewpoint changes present in the HPatches benchmark. Notably, we used the exact same hyperparameters and architectural configuration as our sparse semantic experiments, performing no task-specific tuning for this dense matching objective.

NMT was subsequently evaluated on the HPatches dataset using the Mean Matching Accuracy (MMA) metric at thresholds of 1.0, 3.0, and 5.0 pixels. The results, partitioned into Viewpoint and Illumination subsets, are summarized in Table~\ref{tab:hpatches}.

\begin{table}[!htbp]
\caption{Mean Matching Accuracy (MMA) on the HPatches dataset.}
\centering
\begin{tabular}{lccc}
\toprule
\textbf{Threshold } & \textbf{Viewpoint } & \textbf{Illumination } & \textbf{Overall} \\
\midrule
1.0 pixels & 74.31\% & 91.61\% & 82.81\% \\
3.0 pixels & 75.12\% & 91.61\% & 83.22\% \\
5.0 pixels & 78.09\% & 91.92\% & 84.89\% \\
\bottomrule
\end{tabular}
\label{tab:hpatches}
\end{table}

As shown, despite the lack of specialized hyperparameter tuning, NMT achieves highly competitive accuracy, demonstrating particular robustness to illumination changes ($>91\%$ MMA across all thresholds). However, performance on the viewpoint subset naturally trails specialized dense matchers (such as SuperGlue or LoFTR) which rely heavily on dense homography priors. This confirms that while NMT's normalized architecture is versatile enough to generalize to geometric benchmarks, its primary strength lies in resolving semantic ambiguities rather than pure geometric verification.

\section{Conclusion}
We have introduced a hypersphere‐centric paradigm for sparse keypoint matching, enforcing layer‐wise normalization across all Transformer layers. Coupled with contrastive InfoNCE and hyperspherical uniformity losses, our model learns embeddings that align tightly across images while remaining well-separated within each image. Experiments on PascalVOC and SPair‐71k show state‐of‐the‐art accuracy and require $\geq 1.7\times$ fewer epochs to converge. These results underscore the impact of pervasive normalization and hyperspherical learning, with promising implications for future geometric and structured representation learning tasks.

\bibliographystyle{splncs04}
\bibliography{references}
\end{document}